\documentclass[letterpaper, 10 pt, conference]{ieeeconf}  

\usepackage{times}
\usepackage{graphicx}
\usepackage[ruled,linesnumbered]{algorithm2e}
\usepackage{amsmath}
\usepackage{amsmath}
\usepackage{color,soul}
\usepackage[skip=3pt,font=small,labelfont=bf]{caption}
\usepackage{subcaption}
\usepackage[normalem]{ulem}
\usepackage{multirow}
\usepackage{multicol}
\usepackage{amsfonts}
\usepackage{url}
\usepackage[noadjust]{cite}
\usepackage[bookmarks=true]{hyperref}

\DeclareMathOperator*{\argmin}{arg\,min}

\definecolor{purple}{RGB}{210, 0, 210} 
\definecolor{green}{RGB}{0, 153, 0} 
\newcounter{myenumi}
\renewcommand{\themyenumi}{\arabic{myenumi}.}

\IEEEoverridecommandlockouts                              

\title{\LARGE \bf
Is the Leader Robot an Adequate Sensor for Posture Estimation and Ergonomic Assessment of A Human Teleoperator?}
\author{Amir Yazdani$^{\star}$, Roya Sabbagh Novin$^{\star}$, Andrew Merryweather$^{\star}$, and Tucker Hermans$^{\dagger}$
\thanks{$^{\star}$Department of Mechanical Eng. and Robotics Center, University of Utah, Salt Lake City, UT, USA, 
{\tt\small amir.yazdani@utah.edu} 
}
\thanks{$^{\dagger}$School of Computing and Robotics Center, University of Utah, Salt Lake City, UT, USA, and Nvidia Corporation, Seattle, WA, USA}
}

\begin{document}
\bstctlcite{BSTcontrol}
\maketitle
\thispagestyle{empty}
\pagestyle{empty}
\begin{abstract}
Ergonomic assessment of human posture plays a vital role in understanding work-related safety and health. Current posture estimation approaches face occlusion challenges in teleoperation and physical human-robot interaction. We investigate if the leader robot is an adequate sensor for posture estimation in teleoperation and we introduce a new probabilistic approach that relies solely on the trajectory of the leader robot for generating observations.
We model the human using a redundant, partially-observable dynamical system and we infer the posture using a standard particle filter. We compare our approach with postures from a commercial motion capture system and also two least-squares optimization approaches for human inverse kinematics. The results reveal that the proposed approach successfully estimates human postures and ergonomic risk scores comparable to those estimates from gold-standard motion capture. 
\end{abstract}
\section{Introduction}
\label{sec:introduction}
Work-related musculoskeletal disorders~(WMSDs) are the $2^{\mathrm{nd}}$ largest cause of disabilities worldwide~\cite{vos2015global} and awkward postures are known to contribute to WMSDs. Teleoperation is a well suited alternative for high-risk tasks~(e.g. construction and handling hazardous materials), since the the remote workstation for the human can be designed ergonomically~\cite{dempsey2018emerging}. However, WMSDs are still common among human operators, even when they perform teleoperation without force feedback~\cite{peternel2020human, yu2014ergonomic}. To improve ergonomics and lower the risk of WMSDs in teleoperation, we introduced \textit{Ergonomically Intelligent Teleoperation Systems} in~\cite{yazdani2021posture} and, in this paper, we focus on the problem of posture estimation to assess the ergonomics and risk of WMSDs in teleoperation tasks. Specifically, we investigate if the trajectory information of the leader robot provides adequate sensory information for probabilistic posture estimation for ergonomics assessment.

Posture estimation refers to the process of estimating the kinematic or skeletal configuration of the human body including segment lengths and joint angles. In addition to human biomechanics research, posture estimation is an important part of perception for smart agents~(i.e. collaborative robots~\cite{zanchettin2016safety}, companion mobile robots~\cite{fridovich2020confidence} and self driving cars~\cite{mangalam2020disentangling}) interacting with the humans.
\begin{figure}[t!]
\center
\includegraphics[width = 8.5cm]{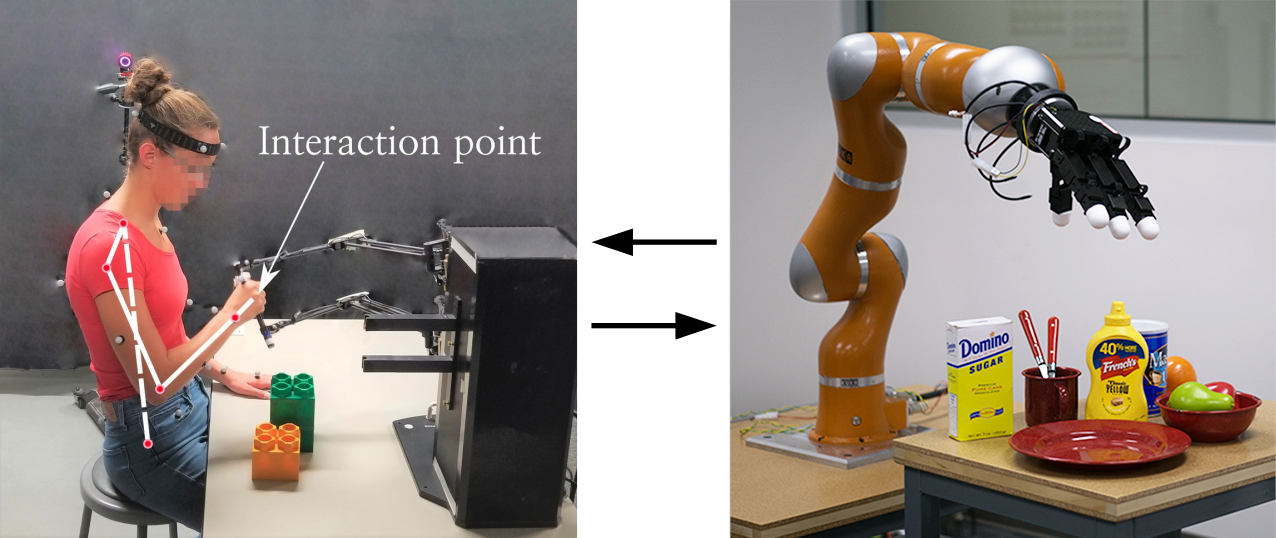}
\caption{Teleoperation setup for human subject experiments including a Quanser HD$^2$ haptic interface. Reflective markers on the participant's body are only used for comparison with a MoCap system.}
\label{fig:setup}
\vspace{-0.7cm}
\end{figure}

Collecting accurate and continuous posture data to determine a risk score over a work shift or entire task cycle can be tedious. Early improvements include using motion capture~(MoCap) systems to estimate the posture and task parameters~(e.g. frequency and duration of the task) still require significant time and effort for set up~\cite{alvarez2017simultaneous}. Moreover, putting markers on human operators can be inconvenient. Alternative markerless techniques are more adaptable, minimally intrusive, and less expensive. However, they need calibration to deal with errors and uncertainties from the sensors~\cite{xiao2018wearable,chen2000camera}. Vision-based markerless methods can also be perturbed by the lighting, background color and even the user's clothing~\cite{erol2007vision}. In teleoperation, a human operator remotely controls the follower robot using the leader robot (see Fig.~\ref{fig:setup}). Using the leader robot in such close proximity to the human increases occlusion and reduces the accurately in estimate posture as mentioned in~\cite{busch2017postural}.

Instead, we propose an alternative non-invasive and probabilistic approach that estimates the posture without using any additional sensors beyond the leader robot necessary for the teleoperation. Our approach could be used either stand-alone for monitoring the human teleoperator posture or in combination with other methods in a multi-modal sensory system to provide robust estimation during occlusions.

In this paper, we formalize posture estimation as a probabilistic inference problem, in which we measure the leader robot's trajectory~(pose and velocity) as the observation and infer the unobserved human posture (joint angles and angular velocities). We use the \textit{circle point analysis}~(CPA)~\cite{mooring1991fundamentals} for segment length estimation of the human body. We also impose physical limits on joint angles and check the validity of the posture based on the posture-dependant ranges of human motion provided by~\cite{jiang2018data}. We incorporate multiple observations over time enabling us to perform inference using a standard particle filter. Then, we use the estimated posture over the course of the task to assess the user's risk of WMSDs using RULA~\cite{McAtamney1993rula}, a standard measure in the ergonomics and safety community. Moreover, we conduct a human subject study and compare the posture estimation results from the particle filter with the results from well-known deterministic solvers for least-squares optimization and the postures from the gold-standard MoCap system. Finally, we show that our posture estimation approach has a good accuracy in risk assessment, comparing to the postures from a MoCap system.
\section{Related Work}
\label{sec:related-work}


In physical human-robot interaction and teleoperation, researchers mainly have used external sensors~(i.e. a vision system or IMU) to estimate a user's posture, especially for hand gestures~\cite{vartholomeos2016design,buzzi2018uncontrolled,martinez}. The idea of solely using the leader robot's trajectory for posture estimation of human teleoperators has been introduced concurrently with this research by Rahal et al. in \cite{rahal2020caring}, where they solved the IK of the 7-DOF human arm. Unlike our probabilistic approach, which can encode a distribution of arm postures, they rely on heuristics to resolve the redundant IK. Their heuristic for redundancy resolution does not always hold across different tasks~(e.g. some tasks might require the human teleoperator to use a working mode different from the working mode of the neutral posture) where the approach in~\cite{rahal2020caring} will fail. 

The application of particle filters in human posture estimation is extensively discussed in the literature~\cite{mozhdehi2018deep, zhang2018correlation, van2001unscented}. The sampling base of particle filters makes them well suited to human posture estimation due to their ability to handle the nonlinearities of human motion~\cite{poppe2007vision}. Moreover, the output estimation is a probabilistic distribution that can preserve different working modes of human posture. However, the high dimension of human motion requires a high number of particles to achieve accurate estimation.

Defining a model for human joint limits is challenging. Studies show that the range of motion~(ROM) for a joint varies depending on the positions of other joints~(inter-joint dependency) or other degrees-of-freedom in the same joint~(intra-joint dependency)~\cite{wang1998three,jiang2018data} and vary by gender and person.
Akhter et al.~\cite{akhter2015pose} used a dataset of recorded MoCap of human motion to develop a discontinous mathematical model for posture-dependant ROM and check the validity of a full-body posture. Jiang et al.~\cite{jiang2018data} used the above model to label the validity of a set of randomly-generated postures and learned a differentiable neural network based on the generated data and used it as a constraint in the inverse kinematics optimization. Their arm model only includes shoulder and elbow and not the wrist. We use this learned network for checking the validity of the arm posture.

The literature highlights that evaluating ergonomics to improve working postures reduces the number of WMSDs~\cite{ismail2010evaluation, khodabakhshi2014ergonomic}. Among all the risk assessment tools, RULA~\cite{McAtamney1993rula} and REBA~\cite{hignett2000rapid} rely mostly on the human posture~(i.e. joint angles) and target the human upper body and whole body, respectively. This makes RULA more suitable for analyzing upper extremity tasks that are common during teleoperation. 
\section{Problem Statement}
\label{sec:problem_statement}
We seek to solve the problem of estimating the human joint-space trajectory in teleoperation using only the observed task-space poses and velocities of the leader robot. We model the physical interaction between the human and the leader robot as an interaction point where the human kinematic chain makes contact with the robot's stylus (Fig.~\ref{fig:setup}).

The state variables of the human include posture~(joint angles) $\mathbf{q}$ and angular velocities $\mathbf{\dot{q}}$. They map into the state variables of the stylus through the kinematics of the human model parameterized by the segment length $\mathbf{\psi}$. We estimate $\mathbf{\psi}$ independently, prior to posture estimation. At each time step, the robot provides an observation as a task-space pose $\mathbf{z}$ and velocity $\mathbf{\dot{z}}$ of the stylus at the interaction point,
\begin{equation}
    [\mathbf{z}_t;\mathbf{\dot{z}}_t]=h(\phi([\mathbf{q}_t;\mathbf{\dot{q}}_t], \mathbf{\psi}))
\end{equation}
where $h$ is the observation function and $\phi$ is the forward kinematics of the human model. This defines only a partial observation of the human posture, because of redundancy in the human kinematics and a noisy measurement at the interaction point, which may change slightly during a task.

We seek to estimate $\boldsymbol{\tau}=[\mathbf{q}_t;\mathbf{\dot{q}}_t]_{t=1:T}$ given the stylus trajectory $\mathcal{Z}=[\mathbf{z}_t;\mathbf{\dot{z}}_t]_{t=1:T}$ that predicts a stylus pose closest to the observed stylus pose and obeys the human motion model $f$~(Eqs.~\ref{eq:vel} and~\ref{eq:pose}):
\begin{align}
\vspace{-12pt}
\mathbf{\tau}^*=&\argmin_{\mathbf{\tau}}\sum_{t=1}^T\!\begin{aligned}[t]
&||\phi([\mathbf{q}_t;\mathbf{\dot{q}}_t], \mathbf{\psi})-[\mathbf{z}_t;\mathbf{\dot{z}}_t]||_{\Sigma_1}^2+\\&||\left[\mathbf{q}_{t},\mathbf{\dot{q}}_{t}\right] - f(\mathbf{q}_{t-1},\mathbf{\dot{q}}_{t-1})||_{\Sigma_2}^2 \label{eq:problem_statement}\end{aligned} \\
                & s.t. \qquad \mathbf{q}_{\mathrm{min}} \leq \mathbf{q} \leq \mathbf{q}_{\mathrm{max}} \nonumber
\end{align}
\begin{figure}[t!]
\vspace{0.2cm}
\hspace{-0.1cm}
\includegraphics[width = 8.8cm]{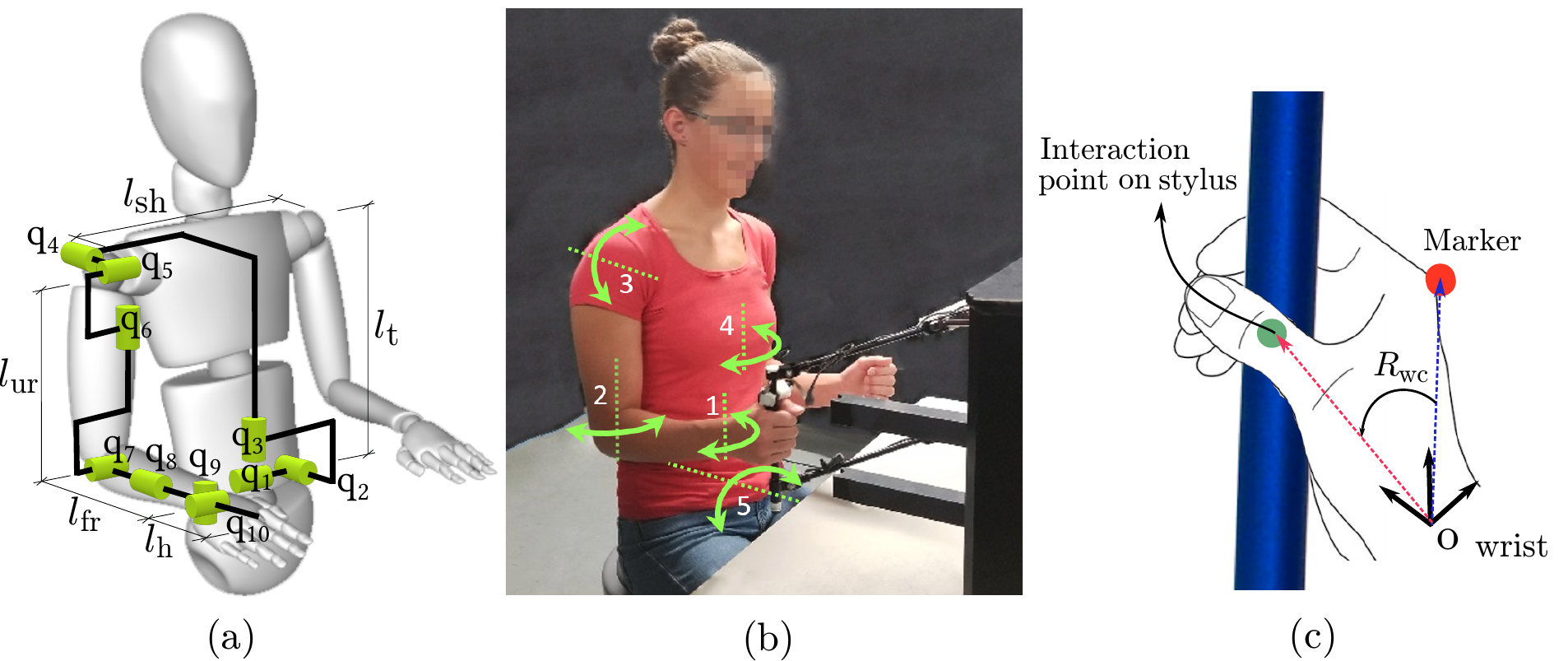}
\caption{(a) Kinematics model of human upper body, (b) Five motion routines for CPA segment length estimation, (c) Hand pose correction for MoCap in grasping the stylus using a fixed rigid-body transformation $R_{wc}$.}
\label{fig:mix_fig}
\vspace{-0.7cm}
\end{figure}where $\mathbf{q}_{\mathrm{min}}$ and $\mathbf{q}_{\mathrm{max}}$ are the joint limits. The high degree-of-freedom in human kinematics makes this problem a \emph{redundant} problem with an infinite number of solutions. We seek the solution closest to the true posture of the human teleoperator.
\section{Approach}
\label{sec:approach}
In this section, we provide an approximate solution for partially observable posture estimation in teleoperation using a particle filter. We provide the kinematics model of the human upper body with only one moving arm. Next, we discuss adopting a particle filter for inference in our problem. Finally, we detail using CPA for segment length estimation.
\subsection{Human kinematics model}
\label{sec:approach-kinematic}
We use a 10-DOF kinematics model (Fig.~\ref{fig:mix_fig}(a)) to analyze the upper body motion of a human sitting on a chair and operating the leader robot. We assume that the chair is stationary with a known position w.r.t. the robot, and the human sits on the center of the chair. The parameters of this model~($\mathbf{\psi}$) include the length of each segment in the upper body model. We compare 3 techniques for segment length estimation: (1) \emph{Full measurement}: manually measuring the segment lengths from anatomical landmarks on participant bodies~\cite{pheasant2018bodyspace, dempster1955space}, (2) \emph{Height measurement}: measuring the height of the participants and selecting the segment lengths fitting to 50-percentile populations from the ANSUR II anthropometric model~(\cite{fromuth2009predicting}), (3) \emph{CPA analysis} where the lengths are calculated from some calibration motion routines described in Sec.~\ref{sec:approach-cpa}.

We define the human's \textit{state variables} as $\mathbf{q}=[q_{i}]_{i =1:10}$, $\mathbf{\dot{q}}=[\dot{q}_{i}]_{i =1:10}$ where $q_i$ represents the angle of joint $i$ (shown in Fig.~\ref{fig:mix_fig}(a)). We assume that the user's hand stays attached to the leader robot's stylus, as such we can transfer the pose of the hand from the human's frame to the robot's frame.

We encode human motion limits in two ways. First, we use fixed limits on the joint angles based on the biomechanics literature~\cite{levangie2000joint, nasa}. 
If an angle estimate exceeds its limits, we project the estimate to the closest limit. Second, we use the learned and posture-dependant joint limit model from \cite{jiang2018data} to ensure the validity of the posture.

The estimated posture of the torso has a high effect on the estimated posture of the arm. Using the full range of motion for the torso would cause challenges in our posture estimation due to the four degrees of redundancy in the kinematics model. To overcome this issue, other researchers assumed that the torso posture is fully known and they only consider the arm~\cite{rahal2020caring}. Instead, we include the torso in the posture estimation problem by assuming that the torso stays close to the vertical position with a low variance. We incorporate this as a perturbance of the torso posture in the problem. This assumption is reasonable and was confirmed in our workstation where the human teleoperator sits behind a table interacting with a haptic interface.

From the kinematics of human motion, we find joint angles and velocities based on the previous step as follows:
\begin{align}
\mathbf{\dot{q}}_k = \mathbf{\dot{q}}_{k-1}+\mathbf{\ddot{q}}_{k-1}dt \label{eq:vel}\\
\mathbf{{q}}_{k} = \mathbf{{q}}_{k-1}+\mathbf{\dot{q}}_{k}dt \label{eq:pose}
\end{align}
We model joint accelerations generated from a Gaussian distribution $\mathbf{\ddot{q}}_{k-1}\sim\mathcal{N}(0,\Tilde{\mathbf{\Sigma}}_v)$. Since $dt$ is fixed, setting \(\mathbf{\Sigma}_v = \Tilde{\mathbf{\Sigma}}_v\cdot dt\) transforms Eq.~(\ref{eq:vel}) to:
\begin{equation}
p(\mathbf{\dot{q}}_k\mid \mathbf{\dot{q}}_{k-1})\sim \mathcal{N}(\mathbf{\dot{q}}_{k-1},\mathbf{\Sigma}_v)
\label{eq:vel_update}
\end{equation}

We model the observation likelihood function by a Gaussian distribution over the hand's pose and velocity as the end-effector of the human kinematic chain:
\begin{equation}
p([\mathbf{z}_k,\mathbf{\dot{z}}_k]\mid [\mathbf{q}_{k},\mathbf{\dot{q}}_{k}]) =\mathcal{N}(\phi(\mathbf{q}_k,\mathbf{\dot{q}}_k,\mathbf{\psi}),\mathbf{\Sigma}_{K})
\label{eq:weight1}
\end{equation}
in which $\mathbf{\Sigma}_K$ is the kinematic covariance matrix. 
\subsection{Particle Filter for Posture Estimation}
\label{sec:approach-pf}
We approximate the solution for the partially observable problem of posture estimation by using a particle filter ~\cite{thrun2005probabilistic} with some modifications. As the estimation of the 10-DOF human model from the trajectory of the leader robot has high ambiguity due to the redundancy, we add the joint angular velocities to our state variables and use the velocity of the leader robot's stylus in our observations. As a prior, we encode that the human \textit{starts} the task in a static, neutral posture as shown in Fig.~\ref{fig:mix_fig}(b). We initialize $M$ particles using a truncated normal distribution with the mean at the neutral posture $\mathbf{q}_{neutral}$ and set the initial angular velocities to zero:
\begin{equation}
{\mathbf{q}}_0^{[m]}\sim \mathcal{N}(\mathbf{q}_{neutral},\Sigma_0),\quad \mathbf{\dot{q}}_{0}^{[m]}=0 \qquad m=1,...,M  \label{eq:process}
\end{equation}
where $\Sigma_0=0.2\times(\mathbf{q}_{\mathrm{max}}-\mathbf{q}_{\mathrm{min}})$ for each joint.

Each particle is propagated in time based on the kinematics of human motion using Eqs.~\ref{eq:vel} and \ref{eq:pose}. Then, the particles are weighted based on the observation likelihood function in Eq.~(\ref{eq:weight1}) defined as the innovation error between the estimated pose of the stylus and the observed pose from the leader robot. We use the multivariate Gaussian distribution to define the likelihood weighting function: 
\begin{align}
w_k^{[m]}=&v_p\cdot \mathrm{det}(2\pi \mathbf{\Sigma}_{K})^{-\frac{1}{2}} \cdot\exp \{-\frac{1}{2}([\mathbf{z}_k,\mathbf{\dot{z}}_k]-\nonumber\\ &\phi(\mathbf{q}_k,\mathbf{\dot{q}}_k,\mathbf{\psi}))^T
\mathbf{\Sigma}_{K}^{-1}([\mathbf{z}_k,\mathbf{\dot{z}}_k]-\phi(\mathbf{q}_k,\mathbf{\dot{q}}_k,\mathbf{\psi}))\}
\label{weight2}
\end{align}
where $v_p \in \mathbb{R}, 0\leq v_p \leq 1$, encodes the validity of the posture as output of the learned neural network from~\cite{jiang2018data}.
\subsection{Circle Point Analysis for Segment Length Estimation}
\label{sec:approach-cpa}
We estimate the segment lengths $\mathbf{\psi}$ through a calibration procedure variant of circle point analysis (CPA)~\cite{mooring1991fundamentals}. Starting from the neutral posture, the user performs five predefined motion patterns that only include motion in one of their joints. When following such a pattern, we assume that the human hand will move on a circle. Estimating the circle parameters defines the location of the active joint and its distance from the end-effector. From this we derive the lengths of each arm segment.

Fig. \ref{fig:mix_fig}(b) presents the five motion patterns used in data generation for CPA: (1)~wrist flexion/extension to estimate hand length; (2)~upper arm external/internal rotation to estimate forearm  length; (3)~upper arm abduction/adduction to estimate upper arm length; (4)~rotation from the hip to estimate shoulder length; and (5)~lateral bending from the hip to estimate torso length. We note that in estimating the last two segments we use the previously estimated arm and hand segment lengths.
\section{Implementation \& Experimental Protocol}
\label{sec:implementation}
\begin{figure}[t!]
\vspace{0.2cm}
\center
\includegraphics[width=\linewidth]{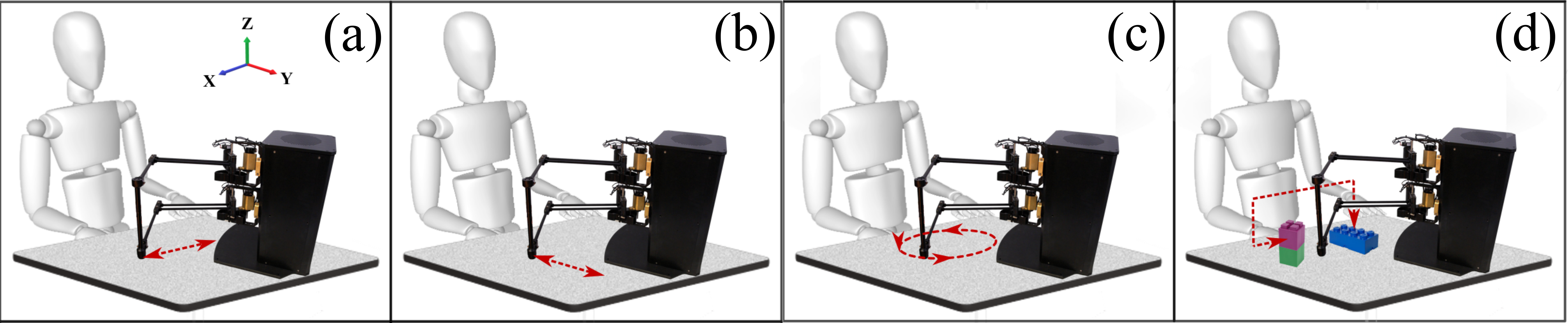}
\caption{The tasks for human subject study: repetitive motions following a straight line in the $X$ direction~(a), a straight line in the $Y$ direction ~(b), a circular path~(c), and repetitive motions between random sides of two blocks positioned at different heights, with an \textit{unprescribed} motion and high range of hand rotation~(d).} \label{fig:exp_tasks}
\vspace{-0.6cm}
\end{figure}

We conducted a human subject experiment in which participants interact with a 6-DOF Quanser HD\(^{2}\) haptic interface as the leader robot~(see Fig.~\ref{fig:setup}), and we recorded their upper body  motion using a 12-camera Optitrack~\cite{optitrack} MoCap system for comparison. We recruited 8 participants~(4 female, 4 male) with ages ranging from 25 to 33 years and heights in the range of $171\pm21$cm. Participants were graduate students from various programs and did not have any experience with teleoperation robots, and each received a 15-min training with the leader robot. Each participant performed 4 tasks visualized in Fig~\ref{fig:exp_tasks}. We provided a printed visual guide on the table for the first three tasks, however, the participants were not required to follow the path accurately. The participants were not told what posture to initialize the task from and how high they should be above the table to do the task. The robot collected data from the participant's motion without exerting any force.

The gold-standard MoCap system estimates the upper-body posture for a 10-DOF torso, however, our human model only includes 3-DOF for the torso. To address this discrepancy, the MoCap posture is retargeted to our human model using the inverse kinematics. Moreover, the segment lengths are variable during a motion in MoCap data, while our model uses fixed lengths. This change is more visible in the forearm and upper arm, where we observed almost 2.3cm and 1.8cm of change, respectively, for a participant doing the circular task. This is mainly because the marker placement on the body will never be perfect, and motion is subject to some skin artifacts leading to this type of error~\cite{metcalf2020quantifying}. Additionally, as shown in Fig.~\ref{fig:mix_fig}(c), the MoCap pose for the hand uses the segment from the wrist axis to the marker at the metacarpophalangeal joint of the index finger, while our human model uses the segment from the wrist joint to the interaction point. To correct for this, we use a fixed rigid-body transformation $R_{wc}$ for the MoCap wrist joint calculated for each participant.

To compare with other well-known approaches, we solved the least-squares problem in Eq.~(\ref{eq:problem_statement}) using two other deterministic methods: (1) boosted and bounded online least-squares IK optimization~(\textit{Online-IK}) in which we simply solve the inverse kinematics optimization independently at each time step by initializing it with the solution from the previous time step, and (2) boosted and bounded offline least-squares trajectory IK optimization~(\textit{Offline-TrajIK}) in which we solve the inverse kinematics problem for the whole trajectory by initializing it with the solution trajectory from the Online-IK. We used dogleg algorithm with rectangular trust regions from SciPy~\cite{virtanen2020scipy} as the optimization solver.

As neither marker-based nor markerless posture estimation techniques provide ground truth posture, we additionally provide qualitative analysis by overlaying the estimated posture on synchronized video frames. Fig.~\ref{fig:frames} shows the posture inferred by our approach aligns well with the MoCap estimates. We see some error due to our fixed segment lengths assumption for MoCap motion. While other possible approaches of analysis exist~(e.g.\ hand-labeling points~\cite{athitsos2003estimating}), these approaches are error prone and time-consuming.

In our implementation, we use a fixed number of particles (M=500), and used $\mathbf{\Sigma}_{v} = 0.01 \cdot \texttt{diag}(0.01,$ $0.01, 0.01, 0.05, 0.05, 0.05)$ and $\mathbf{\Sigma}_{K} = 0.01 \cdot \texttt{diag}(0.001,$ $0.001, 0.001, 0.05, 0.05, 0.05, 1, 1, 1, 10, 10, 10).$
To assess the risk, we developed a code based on RULA which uses both the estimated postures from our approach and the MoCap estimates as the input and outputs the RULA risk score. For that, we used the following assumptions for all tasks: the human is sitting on a chair, minimal intermittent force/load~(less than 2.0Kg), muscle use occurrence less than 4x per minute, untwisted and vertical position for neck and torso, and supported legs and feet.
\section{Results \& Discussion}
\label{sec:results}
This section provides results from our human subject experiments. We discuss the performance of the proposed posture estimation approach comparing with MoCap, as well as the estimated risk assessment results.
\subsection{Segment Lengths Estimation}
\begin{figure}[t!]
\center
\includegraphics[width = 8.7cm]{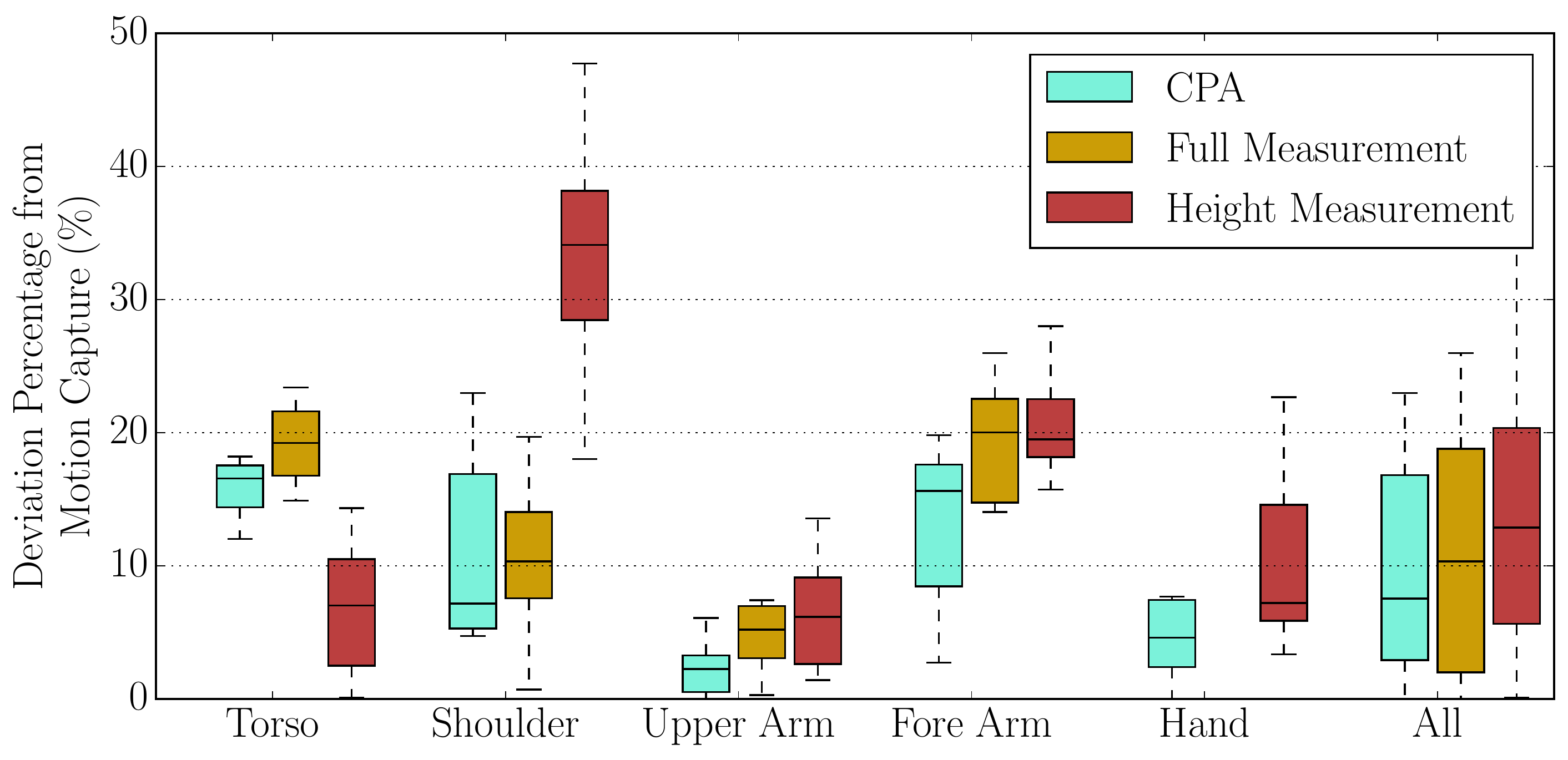}
\caption{Deviation of segments lengths from the MoCap lengths.}
\vspace{-0.2cm}
\label{fig:seg_len_deviation_percent}
\end{figure}

We compared the deviation\footnote{We use the term ``deviation'' instead of ``error'' since the MoCap posture is also an estimate and not ground truth.} of estimated segment lengths from MoCap lengths using the various methods discussed in Section~\ref{sec:approach-kinematic}, among all participants in Fig.~\ref{fig:seg_len_deviation_percent}. The deviation for \emph{full measurement} lengths of the hand is zero since the MoCap marker set did not provide a representative length for the hand. We used the \textit{full measurement} value instead. The last three columns of the figure~(All) shows the deviation for all of the segments. Statistical analyses reveal that \textit{CPA} lengths deviate least from the MoCap lengths significantly\footnote{We use $\alpha=0.05$ for statistical analysis.}. The main reason that \textit{CPA} deviates less than \textit{full measurement} is that the anatomical landmarks used for manual measurement of the lengths are not necessarily on the joint axes calculated by the MoCap from the markers. 
\begin{figure}[t!]
\vspace{0.19cm}
\center
\includegraphics[width=8.7cm]{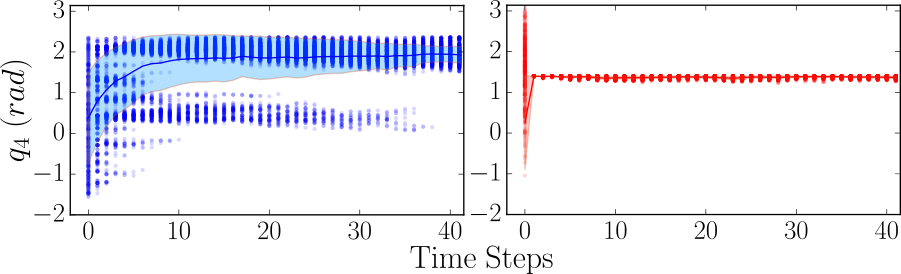}
\caption{Behaviour of particles through time for the shoulder abduction joint of subject~1 during the circular task. (left)~Particles initialized uniformly over the ROM with higher diagonal values of $\Sigma_K$. (right)~Particles initialized from a normal distribution with the mean at the neutral posture and lower diagonal values of $\Sigma_K$.}
\label{fig:converging}
\end{figure}
\begin{figure}[t!]
\center
\includegraphics[width = 8.7cm]{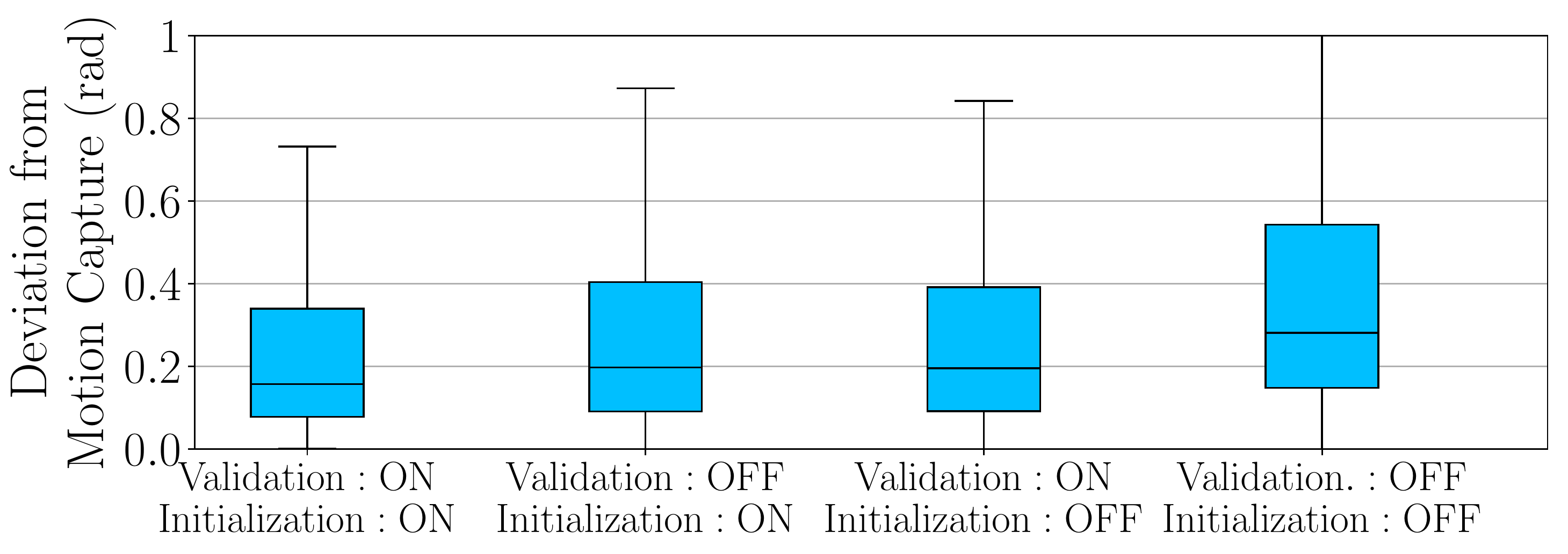}
\caption{The effect of initialization and posture validation on posture estimation in trials of the circular and the horizontal tasks.}
\vspace{-0.4cm}
\label{fig:posture_validity}
\end{figure}

\subsection{Initialization and Validation of Particles}
Fig.~\ref{fig:converging} shows the evolution of all the particles in a trial of task~(3) with circular motion, as well as the mean and standard deviation of their distributions. On the left, we initialized particles from a uniform distribution over a joint's ROM and used high values of kinematic covariance~(0.1 for position, 0.5 for orientation). Here, particles from multiple modes are kept for about 40 steps, then they converge into a single mode. On the right, we initialize particles from a normal distribution around the neutral posture and the kinematic covariance matrix~$\Sigma_0$ as described in the Section~\ref{sec:approach-kinematic}.
The particles converge quickly to the correct mode. This shows the effect of initializing the particles around the neutral posture. Although we did not suggest the any initial posture to the participants, we observed that they all started from a posture close to their neutral posture independent of the type of the task they perform, which supports our idea of initialization of particles around the neutral posture.

Moreover, Fig.~\ref{fig:posture_validity} shows the comparison between the effect of initialization around the neutral posture and validation of the posture based on the posture-dependant ROM on the deviation of the estimated postures from the MoCap postures. This is done for 4 trials of the task~(1) and 4 trials of the task~(3). We see that the combination of these two methods reduces the deviation, however, the effect of each one is almost equal. This implies that in some applications, it is safe to just use the initialization around the neutral posture instead of the computationally-expensive validation of particles.

\subsection{Posture Estimation}
\begin{figure}[t!]
\begin{centering}
\includegraphics[width=8.7cm]{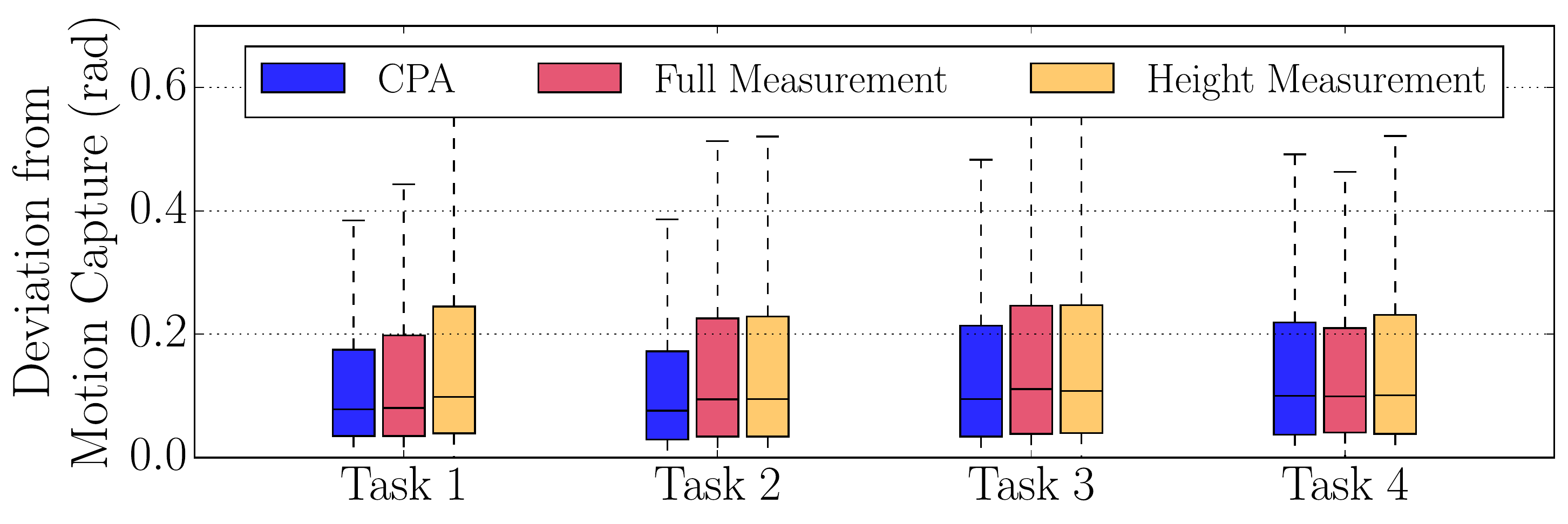}
\caption{Deviation of the posture estimated by the proposed approach from MoCap system for all the participants in different tasks.}
\label{fig:boxplot_tasks}
\vspace{-0.2cm}
\end{centering}
\end{figure}

Fig.~\ref{fig:boxplot_tasks} represents the deviation between the posture from our approach and the posture from MoCap for all the participants, and trials across the tasks using different segment length estimation methods. Overall, the approach generally agrees with a median deviation less than 0.09rad~(less than 5deg) and upper quartile less than 0.25rad~(less than 15deg) considering the observation solely from the stylus trajectory and having no extra sensors.
Statistical analysis for comparing the effect of all three segment length estimation methods on the posture estimation accuracy reveals that the estimated postures using \textit{CPA} method has a significantly lower deviation from the MoCap postures, where there is no significant difference between the \textit{full measurement} and \textit{height measurement} methods. Hence, we used segment lengths from \textit{CPA} method for the rest of the experiments.
\begin{figure}[t!]
\begin{centering}
\includegraphics[width=8.7cm]{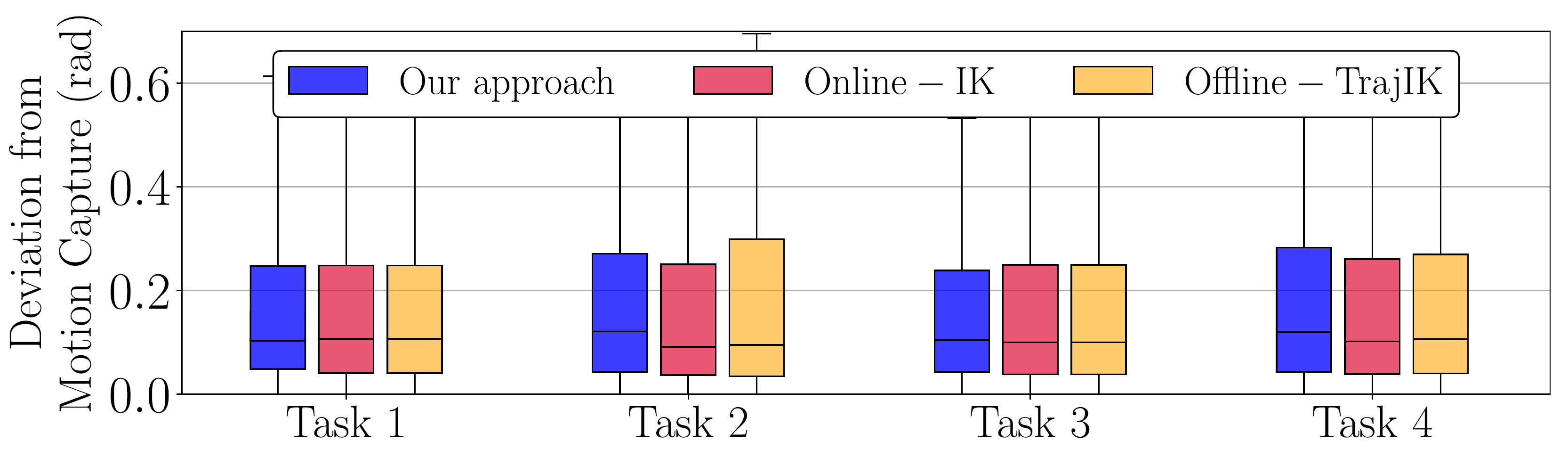}
\caption{Deviation of the posture estimated by our approach vs Online-IK and Offline-TrajIK methods among all the tasks.}
\label{fig:ik_methods_tasks}
\vspace{-0.4cm}
\end{centering}
\end{figure}

Fig.~\ref{fig:ik_methods_tasks} compares our posture estimation approach with two other least-squares IK solutions for redundant robots, for all participants among 4 tasks. From the statistical analysis of the results, we can conclude that our probabilistic \textit{particle filter} approach performance is not significantly different than the other two methods. 

\begin{figure*}[t!]
\begin{centering}
\includegraphics[width=17.7cm]{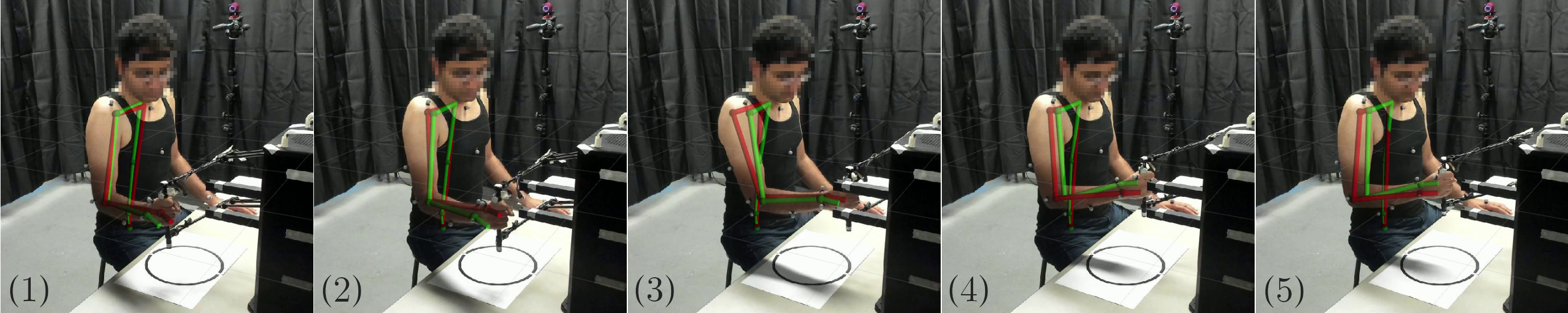}
\caption{Video-overlaid skeletons show the posture~(most probable particle) estimated from our proposed approach~(red) and MoCap~(green).}
\label{fig:frames}
\end{centering}
\vspace{-0.5cm}
\end{figure*}

Since there is no ground truth for human posture to compare with, we provide a qualitative evaluation using video frames of a participant during task~(3) with overlaid reconstructed skeletons from MoCap~(green) and our approach~(red) in Fig.~\ref{fig:frames}. The estimated postures aligns well with the MoCap postures during the entire task. Videos of overlaid skeletons for all four tasks are available in the supplementary video. Although the leader robot's trajectory is smooth, the videos show non-smooth estimations from our approach, which is due to the characteristics of the particle filter and plotting the skeleton for the most-probable posture. 

\subsection{Risk Assessment}
\begin{figure}[t!]
\begin{centering}
\includegraphics[width=8.7cm]{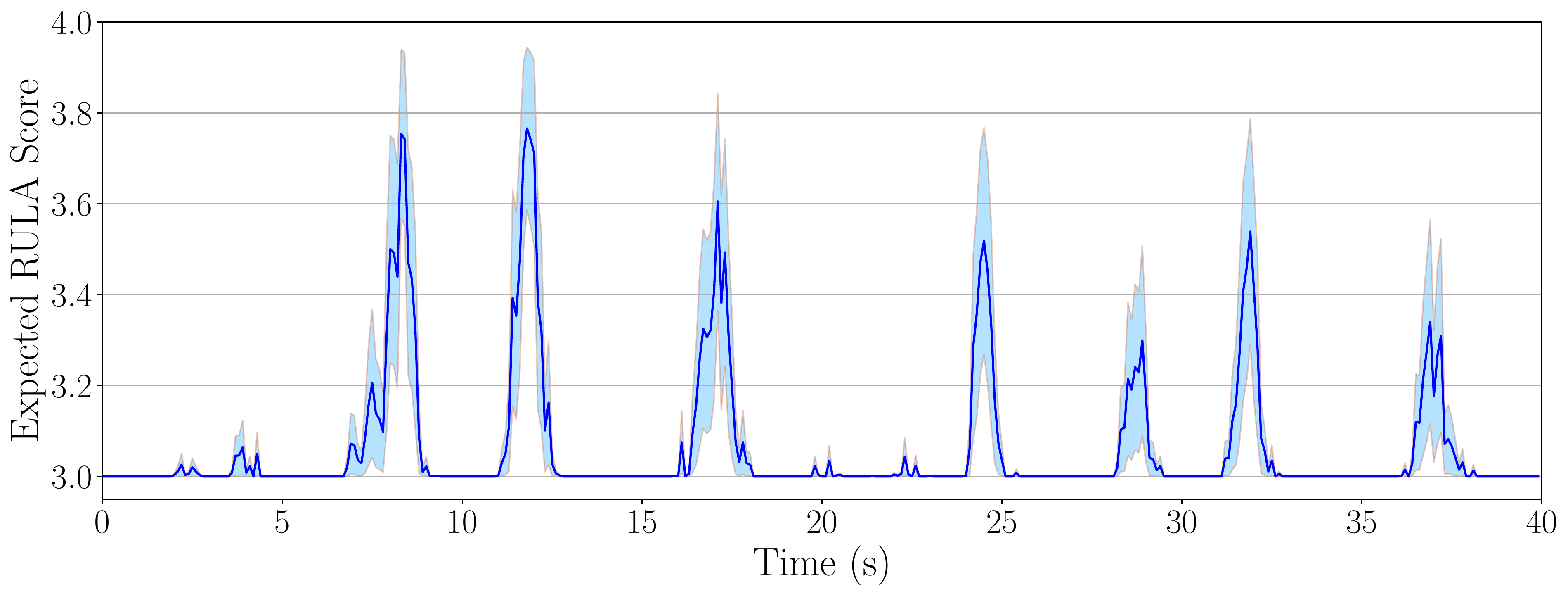}
\caption{Expected RULA score (solid line) and standard deviation
  (shaded region) over time for participant 1 performing task~(1).}
\label{fig:expected_rula}
\end{centering}
\end{figure}

To represent the benefits of our probabilistic approach, we plot the expected value of RULA score and its standard deviation over time during task~(1) using the posture estimates from our particle filter in Fig.~\mbox{\ref{fig:expected_rula}}. Unlike the deterministic estimators, our probabilistic approach provides an estimated distribution over posture and hence, a distribution over estimated RULA scores which can be used by probabilistic human-aware planning methods in p-HRI and teleoperation.

\begin{figure}[t!]
\begin{centering}
\includegraphics[width=8.5cm]{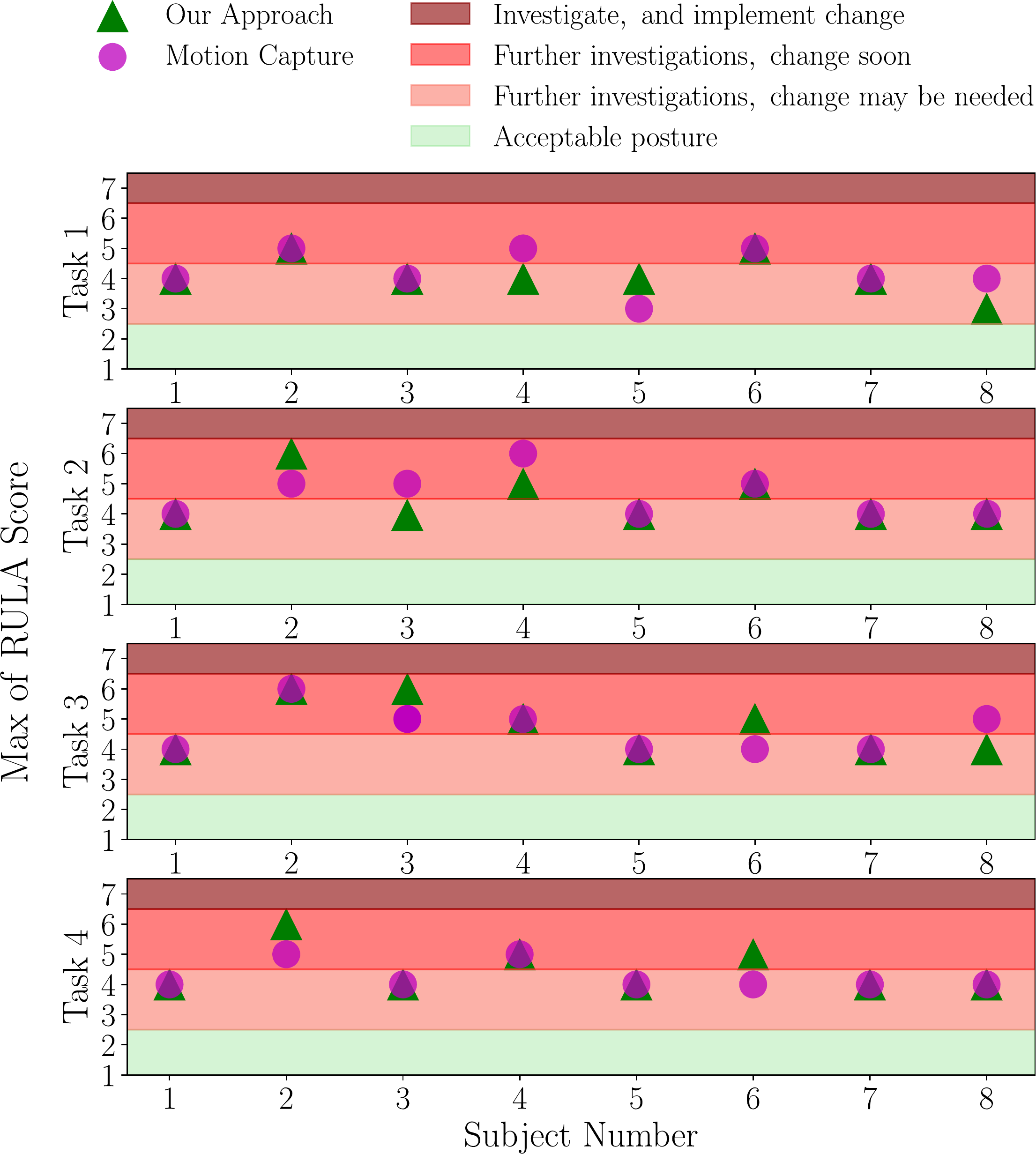}
\caption{Comparison of the maximum value of RULA scores of a task~(using estimated posture from our approach and MoCap for all the participants and trials.}
\label{fig:rula_subject}
\end{centering}
\vspace{-0.6cm}
\end{figure}

Fig.~\ref{fig:rula_subject} shows the maximum RULA score during a task~(the one most often used in ergonomics) for the postures from our approach and MoCap postures. Our approach was successful in identifying all instances where the RULA score was higher than 2~(i.e. future investigation or change may be or is needed) in all 32 trials. The experiment resulted in the same interpretation of the RULA score in 27 trials~(84.37\%) and the same RULA score in 21 trials~(65.63\%). Our approach estimates the same RULA score~(not the maximum value for task) across all trajectories and participants with median accuracy of more than 86.4\% for tasks~(1) and (2), and more than 74.7\% for tasks~(3) and (4).

Overall, the results show that the proposed posture estimation solely
from the leader robot has the potential to be used for continuous
monitoring of ergonomics in teleoperation. It is also
accurate enough to provide alerts when further ergonomics
investigation or change in the task is required.

\section{Conclusion}
\label{sec:conclusion}
In this paper, we investigated if a leader robot is an adequate sensor to continuously monitor posture to assess the ergonomics of a human-teleoperated task. We described a probabilistic approach which is based solely on the data recorded from a leader robot's end-effector, that is already necessary to perform the teleoperation task. We compared our approach with well-known \textit{Online-IK} and \textit{Offline-TrajIK} methods for human inverse kinematics. We used CPA to estimate segment lengths and utilized RULA for risk assessment.

Our results show that the leader robot can be an adequate sensor for human posture estimation and ergonomic assessment in applications such as continuous monitoring of the human teleoperator, while our assumptions are valid. More specifically, we found that \textit{CPA} estimates more accurate segment lengths and the proposed algorithm can successfully estimate the posture based solely on the leader robot's stylus trajectory with a low deviation from MoCap. We also showed that our approach agrees with MoCap postures similar to \textit{Online-IK} and \textit{Offline-TrajIK}, providing a probabilistic distribution for the posture. Furthermore, the risk assessment results show that our proposed approach resulted in the same interpretation of the RULA score in 27 of 32 trials~(84.37\%), and the same maximum RULA score in 21 trials~(65.63\%). This is sufficient to trigger further assessment to investigate ergonomic hazards.

In this paper, we assumed a seated human teleoperator on a chair with known relative position to the leader robot, but we can extend our approach to a standing human by adding the foot pose as a state variable. Here, we focused on teleoperation applications, which has a high prevalence of WMSDs, However, it is possible to extend it to other physical human-robot interaction tasks such as co-manipulation and learning from demonstration by a human expert, in order to assess the risk of WMSDS and also to enhance the learning performance by including the human's posture and comfort during demonstration. Our approach also can be used for full arm representation in VR systems based on the pose of controllers and headset. It would be straightforward to combine our proposed approach with other sensing modalities (e.g., vision or MoCap) when available in a specific application context by integrating these additional observations into the particle filter weighting scheme.

Future work will focus on increasing the complexity of the human model~(e.g. adding muscle activation). Additionally, we are examining substituting the particle filter with an incremental smoothing method~\cite{kaess2012isam2} in order to improve estimation results and decrease runtime. 

\newpage
\bibliographystyle{IEEEtran}
\bibliography{references}

\begin{thebibliography}{10}
\providecommand{\url}[1]{#1}
\csname url@rmstyle\endcsname
\providecommand{\newblock}{\relax}
\providecommand{\bibinfo}[2]{#2}
\providecommand\BIBentrySTDinterwordspacing{\spaceskip=0pt\relax}
\providecommand\BIBentryALTinterwordstretchfactor{4}
\providecommand\BIBentryALTinterwordspacing{\spaceskip=\fontdimen2\font plus
\BIBentryALTinterwordstretchfactor\fontdimen3\font minus
  \fontdimen4\font\relax}
\providecommand\BIBforeignlanguage[2]{{%
\expandafter\ifx\csname l@#1\endcsname\relax
\typeout{** WARNING: IEEEtran.bst: No hyphenation pattern has been}%
\typeout{** loaded for the language `#1'. Using the pattern for}%
\typeout{** the default language instead.}%
\else
\language=\csname l@#1\endcsname
\fi
#2}}

\bibitem{vos2015global}
T.~Vos, R.~M. Barber, B.~Bell, A.~Bertozzi-Villa, S.~Biryukov, I.~Bolliger,
  F.~Charlson, A.~Davis, L.~Degenhardt, D.~Dicker, \emph{et~al.},
  ``\href{https://www.sciencedirect.com/science/article/pii/S0140673615606924}{Global,
  regional, and national incidence, prevalence, and years lived with disability
  for 301 acute and chronic diseases and injuries in 188 countries, 1990--2013:
  a systematic analysis for the Global Burden of Disease Study 2013},''
  \emph{The Lancet}, vol. 386, no. 9995, pp. 743--800, 2015.

\bibitem{dempsey2018emerging}
P.~G. Dempsey, L.~M. Kocher, M.~F. Nasarwanji, J.~P. Pollard, and A.~E.
  Whitson, ``\href{https://www.mdpi.com/1660-4601/15/11/2449}{Emerging
  ergonomics issues and opportunities in mining},'' \emph{International journal
  of environmental research and public health}, vol.~15, no.~11, p. 2449, 2018.

\bibitem{peternel2020human}
L.~Peternel, C.~Fang, M.~Laghi, A.~Bicchi, N.~Tsagarakis, and A.~Ajoudani,
  ``Human arm posture optimisation in bilateral teleoperation through interface
  reconfiguration,'' in \emph{IEEE RAS/EMBS International Conference for
  Biomedical Robotics and Biomechatronics (BioRob 2020)}, 2020.

\bibitem{yu2014ergonomic}
S.~Yu, J.~Lee, B.~Park, K.~Kim, and I.~Cho,
  ``\href{https://www.sciencedirect.com/science/article/pii/S1738573315301170}{Ergonomic
  analysis of a telemanipulation technique for a pyroprocess demonstration
  facility},'' \emph{Nuclear Engineering and Technology}, vol.~46, no.~4, pp.
  489--500, 2014.

\bibitem{yazdani2021posture}
A.~Yazdani and R.~Sabbagh~Novin,
  ``\href{https://dl.acm.org/doi/abs/10.1145/3434074.3446350}{Posture
  Estimation and Optimization in Ergonomically Intelligent Teleoperation
  Systems},'' in \emph{Companion of the 2021 ACM/IEEE International Conference
  on Human-Robot Interaction}, 2021, pp. 604--606.

\bibitem{zanchettin2016safety}
A.~M. Zanchettin, N.~M. Ceriani, P.~Rocco, H.~Ding, and B.~Matthias,
  ``\href{https://ieeexplore.ieee.org/abstract/document/7079531}{Safety in
  human-robot collaborative manufacturing environments: Metrics and control},''
  \emph{IEEE Transactions on Automation Science and Engineering}, vol.~13,
  no.~2, pp. 882--893, 2016.

\bibitem{fridovich2020confidence}
D.~Fridovich-Keil, A.~Bajcsy, J.~F. Fisac, S.~L. Herbert, S.~Wang, A.~D.
  Dragan, and C.~J. Tomlin,
  ``\href{https://journals.sagepub.com/doi/full/10.1177/0278364919859436}{Confidence-aware
  motion prediction for real-time collision avoidance1},'' \emph{The
  International Journal of Robotics Research}, vol.~39, no. 2-3, pp. 250--265,
  2020.

\bibitem{mangalam2020disentangling}
K.~Mangalam, E.~Adeli, K.-H. Lee, A.~Gaidon, and J.~C. Niebles,
  ``\href{https://openaccess.thecvf.com/content_WACV_2020/html/Mangalam_Disentangling_Human_Dynamics_for_Pedestrian_Locomotion_Forecasting_with_Noisy_Supervision_WACV_2020_paper.html}{Disentangling
  human dynamics for pedestrian locomotion forecasting with noisy
  supervision},'' in \emph{The IEEE Winter Conference on Applications of
  Computer Vision}, 2020, pp. 2784--2793.

\bibitem{alvarez2017simultaneous}
M.~Alvarez, D.~Torricelli, A.~J. del Ama, D.~Pinto, J.~Gonzalez-Vargas, J.~C.
  Moreno, A.~Gil-Agudo, and J.~L. Pons,
  ``\href{https://ieeexplore.ieee.org/abstract/document/8009449/}{Simultaneous
  estimation of human and exoskeleton motion: a simplified protocol},'' in
  \emph{IEEE Int. Conf. on Rehabilitation Robotics}, 2017, pp. 1431--1436.

\bibitem{xiao2018wearable}
X.~Xiao and S.~Zarar,
  ``\href{https://ieeexplore.ieee.org/abstract/document/8487858}{A wearable
  system for articulated human pose tracking under uncertainty of sensor
  placement},'' in \emph{IEEE Int. Conf. on Biomedical Robotics and
  Biomechatronics}, 2018, pp. 1144--1150.

\bibitem{chen2000camera}
X.~Chen and J.~Davis,
  ``\href{http://graphics.stanford.edu/papers/OcclusionMetric/occlusion_metric.pdf}{Camera
  placement considering occlusion for robust motion capture},'' Computer
  Graphics Laboratory, Stanford University, Tech. Rep, Tech. Rep., 2000.

\bibitem{erol2007vision}
A.~Erol, G.~Bebis, M.~Nicolescu, R.~D. Boyle, and X.~Twombly,
  ``\href{https://www.sciencedirect.com/science/article/pii/S1077314206002281}{Vision-based
  hand pose estimation: A review},'' \emph{Computer Vision and Image
  Understanding}, vol. 108, no. 1-2, pp. 52--73, 2007.

\bibitem{busch2017postural}
B.~Busch, G.~Maeda, Y.~Mollard, M.~Demangeat, and M.~Lopes,
  ``\href{https://ieeexplore.ieee.org/abstract/document/8206107}{Postural
  optimization for an ergonomic human-robot interaction},'' in \emph{2017
  IEEE/RSJ International Conference on Intelligent Robots and Systems
  (IROS)}.\hskip 1em plus 0.5em minus 0.4em\relax IEEE, 2017, pp. 2778--2785.

\bibitem{mooring1991fundamentals}
B.~W. Mooring, Z.~S. Roth, and M.~R. Driels,
  \emph{\href{https://dl.acm.org/doi/book/10.5555/574653}{Fundamentals of
  manipulator calibration}}.\hskip 1em plus 0.5em minus 0.4em\relax Wiley New
  York, 1991.

\bibitem{jiang2018data}
Y.~Jiang and C.~K. Liu,
  ``\href{https://ieeexplore.ieee.org/abstract/document/8461010}{Data-driven
  approach to simulating realistic human joint constraints},'' in \emph{2018
  IEEE International Conference on Robotics and Automation (ICRA)}.\hskip 1em
  plus 0.5em minus 0.4em\relax IEEE, 2018, pp. 1098--1103.

\bibitem{McAtamney1993rula}
L.~McAtamney and E.~N. Corlett,
  ``\href{https://www.sciencedirect.com/science/article/abs/pii/000368709390080S}{RULA:
  a survey method for the investigation of work-related upper limb
  disorders},'' \emph{Applied ergonomics}, vol.~24, no.~2, pp. 91--99, 1993.

\bibitem{vartholomeos2016design}
P.~Vartholomeos, N.~Katevas, A.~Papadakis, and L.~Sarakis,
  ``\href{https://ieeexplore.ieee.org/abstract/document/7551933}{Design of
  motion-tracking device for intuitive and safe human-robot physical
  interaction},'' in \emph{IEEE Int. Conf. on Telecommunications and
  Multimedia}, 2016, pp. 1--6.

\bibitem{buzzi2018uncontrolled}
\BIBentryALTinterwordspacing
J.~Buzzi, E.~De~Momi, and I.~Nisky,
  ``\href{https://ieeexplore.ieee.org/abstract/document/8370066}{An
  uncontrolled manifold analysis of arm joint variability in virtual planar
  position and orientation tele-manipulation},'' \emph{IEEE Transactions on
  Biomedical Engineering}, 2018.
\BIBentrySTDinterwordspacing

\bibitem{martinez}
G.~Martinez, I.~A. Kakadiaris, D.~Magruder, and I.~Magruder,
  ``\href{https://pdfs.semanticscholar.org/c271/2f94d0d836faf922b796567ab3c8ccfe4409.pdf}{{Teleoperating
  ROBONAUT: A case study}},'' in \emph{British Machine Vision Conference},
  2002, pp. 1--10.

\bibitem{rahal2020caring}
R.~Rahal, G.~Matarese, M.~Gabiccini, A.~Artoni, D.~Prattichizzo, P.~R.
  Giordano, and C.~Pacchierotti,
  ``\href{https://ieeexplore.ieee.org/abstract/document/8970329/}{Caring about
  the human operator: haptic shared control for enhanced user comfort in
  robotic telemanipulation},'' \emph{IEEE Transactions on Haptics}, vol.~13,
  no.~1, pp. 197--203, 2020.

\bibitem{mozhdehi2018deep}
R.~J. Mozhdehi, Y.~Reznichenko, A.~Siddique, and H.~Medeiros,
  ``\href{https://ieeexplore.ieee.org/abstract/document/8451069}{Deep
  convolutional particle filter with adaptive correlation maps for visual
  tracking},'' in \emph{IEEE Int. Conf. on Image Processing}, 2018, pp.
  798--802.

\bibitem{zhang2018correlation}
T.~Zhang, S.~Liu, C.~Xu, B.~Liu, and M.~Yang,
  ``\href{https://ieeexplore.ieee.org/abstract/document/8170277}{Correlation
  particle filter for visual tracking},'' \emph{IEEE Transactions on Image
  Processing}, vol.~27, no.~6, pp. 2676--2687, 2018.

\bibitem{van2001unscented}
R.~Van Der~Merwe, A.~Doucet, N.~De~Freitas, and E.~A. Wan,
  ``\href{http://papers.nips.cc/paper/1818-the-unscented-particle-filter.pdf}{The
  Unscented Particle Filter},'' in \emph{Advances in neural information
  processing systems}, 2001, pp. 584--590.

\bibitem{poppe2007vision}
R.~Poppe,
  ``\href{https://www.sciencedirect.com/science/article/pii/S1077314206002293}{Vision-based
  human motion analysis: An overview},'' \emph{Computer vision and image
  understanding}, vol. 108, no. 1-2, pp. 4--18, 2007.

\bibitem{wang1998three}
X.~Wang, M.~Maurin, F.~Mazet, N.~D.~C. Maia, K.~Voinot, J.~P. Verriest, and
  M.~Fayet,
  ``\href{https://www.sciencedirect.com/science/article/pii/S0021929098000980}{Three-dimensional
  modelling of the motion range of axial rotation of the upper arm},''
  \emph{Journal of biomechanics}, vol.~31, no.~10, pp. 899--908, 1998.

\bibitem{akhter2015pose}
I.~Akhter and M.~J. Black,
  ``\href{https://ieeexplore.ieee.org/document/7298751}{Pose-conditioned joint
  angle limits for 3D human pose reconstruction},'' in \emph{Proceedings of the
  IEEE conference on computer vision and pattern recognition}, 2015, pp.
  1446--1455.

\bibitem{ismail2010evaluation}
S.~A. Ismail, S.~B.~M. Tamrin, M.~R. Baharudin, M.~A.~M. Noor, M.~H. Juni,
  J.~Jalaludin, and Z.~Hashim,
  ``\href{http://docsdrive.com/pdfs/medwelljournals/rjmsci/2010/1-10.pdf}{Evaluation
  of two ergonomics intervention programs in reducing ergonomic risk factors of
  musculoskeletal disorder among school children},'' \emph{Res J Med Sci},
  vol.~4, no.~1, pp. 1--10, 2010.

\bibitem{khodabakhshi2014ergonomic}
Z.~Khodabakhshi, S.~A. Saadatmand, M.~Anbarian, and R.~Heydari~Moghadam, ``An
  ergonomic assessment of musculoskeletal disorders risk among the computer
  users by rula technique and effects of an eight-week corrective exercises
  program on reduction of musculoskeletal pain,'' \emph{Iranian Journal of
  Ergonomics}, vol.~2, no.~3, pp. 44--56, 2014.

\bibitem{hignett2000rapid}
S.~Hignett and L.~McAtamney,
  ``\href{https://www.sciencedirect.com/science/article/pii/S0003687099000393}{Rapid
  entire body assessment (REBA)},'' \emph{Applied ergonomics}, vol.~31, no.~2,
  pp. 201--205, 2000.

\bibitem{pheasant2018bodyspace}
S.~Pheasant and C.~M. Haslegrave, \emph{Bodyspace: Anthropometry, ergonomics
  and the design of work}.\hskip 1em plus 0.5em minus 0.4em\relax CRC press,
  2018.

\bibitem{dempster1955space}
W.~T. Dempster, ``Space requirements of the seated operator, geometrical,
  kinematic, and mechanical aspects of the body with special reference to the
  limbs,'' Michigan State Univ East Lansing, Tech. Rep., 1955.

\bibitem{fromuth2009predicting}
R.~C. Fromuth and M.~B. Parkinson,
  ``\href{https://asmedigitalcollection.asme.org/IDETC-CIE/proceedings-abstract/IDETC-CIE2008/43253/581/330839}{Predicting
  5th and 95th percentile anthropometric segment lengths from population
  stature},'' in \emph{ASME International Design Engineering Technical
  Conferences and Computers and Information in Engineering}, 2009, pp.
  581--588.

\bibitem{levangie2000joint}
P.~K. Levangie and C.~C. Norkin, ``Joint structure and function; a
  comprehensive analysis. 3rd,'' \emph{Philadelphia: FA. Davis Company}, 2000.

\bibitem{nasa}
``{NASA} man-system integration standard,'' Available at
  \url{https://msis.jsc.nasa.gov/sections/section03.htm} (09/16/2020).

\bibitem{thrun2005probabilistic}
S.~Thrun, W.~Burgard, and D.~Fox,
  \emph{\href{http://www.probabilistic-robotics.org/}{Probabilistic
  robotics}}.\hskip 1em plus 0.5em minus 0.4em\relax MIT press, 2005.

\bibitem{optitrack}
``{NaturalPoint Inc., Corvalis, OR},'' \url{https://optitrack.com/}
  (09/16/2020).

\bibitem{metcalf2020quantifying}
C.~Metcalf, C.~Phillips, A.~Forrester, J.~Glodowski, K.~Simpson, C.~Everitt,
  A.~Darekar, L.~King, D.~Warwick, and A.~Dickinson,
  ``\href{https://link.springer.com/article/10.1007/s10439-020-02476-2}{Quantifying
  Soft Tissue Artefacts and Imaging Variability in Motion Capture of the
  Fingers},'' \emph{Annals of Biomedical Engineering}, pp. 1--11, 2020.

\bibitem{virtanen2020scipy}
P.~Virtanen, R.~Gommers, T.~E. Oliphant, M.~Haberland, T.~Reddy, D.~Cournapeau,
  E.~Burovski, P.~Peterson, W.~Weckesser, J.~Bright, \emph{et~al.},
  ``\href{https://www.nature.com/articles/s41592-019-0686-2}{SciPy 1.0:
  fundamental algorithms for scientific computing in Python},'' \emph{Nature
  methods}, vol.~17, no.~3, pp. 261--272, 2020.

\bibitem{athitsos2003estimating}
V.~Athitsos and S.~Sclaroff,
  ``\href{https://ieeexplore.ieee.org/abstract/document/1211500/}{Estimating 3D
  hand pose from a cluttered image},'' in \emph{IEEE Conference on Computer
  Vision and Pattern Recognition}, vol.~2, 2003, pp. II--432.

\bibitem{kaess2012isam2}
\BIBentryALTinterwordspacing
M.~Kaess, H.~Johannsson, R.~Roberts, V.~Ila, J.~J. Leonard, and F.~Dellaert,
  ``\href{https://journals.sagepub.com/doi/abs/10.1177/0278364911430419}{iSAM2:
  Incremental smoothing and mapping using the Bayes tree},'' \emph{The
  International Journal of Robotics Research}, vol.~31, no.~2, pp. 216--235,
  2012.
\BIBentrySTDinterwordspacing

\end{thebibliography}

\end{document}